# An Investigation of Smart Contract for Collaborative Machine Learning Model Training

Shengwen Ding

kamisama.ding@gmail.com

Chenhui Hu

chenhui.hu@gmail.com

*Abstract*—Machine learning (ML) has penetrated various fields in the era of big data. The advantage of collaborative machine learning (CML) over most conventional ML lies in the joint effort of decentralized nodes or agents that results in better model performance and generalization. As the training of ML models requires a massive amount of good quality data, it is necessary to eliminate concerns about data privacy and ensure high-quality data. To solve this problem, we cast our eyes on the integration of CML and smart contracts. Based on blockchain, smart contracts enable automatic execution of data preserving and validation, as well as the continuity of CML model training. In our simulation experiments, we define incentive mechanisms on the smart contract, investigate the important factors such as the number of features in the dataset (num_words), the size of the training data, the cost for the data holders to submit data, etc., and conclude how these factors impact the performance metrics of the model: the accuracy of the trained model, the gap between the accuracies of the model before and after simulation, and the time to use up the balance of bad agent. For instance, the increase of the value of num_words leads to higher model accuracy and eliminates the negative influence of malicious agents in a shorter time from our observation of the experiment results. Statistical analyses show that with the help of smart contracts, the influence of invalid data is efficiently diminished and model robustness is maintained. We also discuss the gap in existing research and put forward possible future directions for further works.

## I. Introduction

Machine learning (ML), a field that improves computer algorithms through the learning experience, has been widely applied in various fields so far such as lip reading [1], image classification [2], speech recognition [3], genome sequence analysis [4], cancer prediction [5], etc. Compared with conventional ML model training that is generally centralized, the difference of collaborative machine learning (CML) model training lies in the joint effort of decentralized nodes (e.g. compute clusters, GPUs) or agents (e.g. edge device users). The data is usually distributed across many agents or nodes, instead of one single agent or server. Therefore, in the era of big data, CML is a natural choice that results in better model performance, i.e. improving generalization and robustness to label noise [6].

Since it requires a massive amount of good quality data (so-called big data) for ML model training before generalizing into real-life tasks [7], it is necessary to eliminate concerns about data privacy. Blockchain [8], a peer-to-peer decentralized ledger for recording transaction data, is useful to encourage secured collaborative data sharing attributed to its properties of transparency, traceability, and immutability. For instance, the DInEMMo framework [29], a marketplace with incentivized mechanism automated by blockchain, can be an efficient platform for enhancing ML models in the medical field through the collaborative sharing of medical diagnostic data from various hospitals. On the concern of high-quality data and continuous training, smart contract [9], a protocol logically written up in codes based on blockchain, is able to realize automatic execution of the filtering of bad data as well as sustainable model training.

In this investigation, we have done extensive experiemtns to identify the most important parameters on the performance of the model to inform reference for further study. Our main focus is to incentive collaborative data sharing for ML model training through the utilization of blockchain. A perceptron model is trained on the IMDB reviews dataset [10] for sentiment classification in our simulation experiments which emphasizes important factors of the smart contract such as the number of features of the dataset (num_words), the size of the training data (train_size), the mean of the deposit required to submit data (mean_deposit), and the cost deducted from the deposit for each submission of new data (submission cost). The major findings from the research results on the most important factors are as follows:

- num_words: The model accuracies tend to increase with the growth of num_words. When the number of features in the dataset is large enough (800 in this case), the model can learn useful features from the complex patterns in the data and the incentive mechanism can screen bad data more correctly.

- train_size: When a too-small proportion of all data is used to train the model before simulation, the model accuracy is unstable, the time spent to use up the balance of the malicious agent is longer, and the result has little reference value. The model accuracies tend to increase with the amount of train_size. When the train_size is not too small (for example not below 5%), the accuracy of the model before the simulation can maintain stability and has a referential value.

- mean_deposit: In general, the mean_deposit does not have as significant an impact on the overall model performance as the previous parameters. The effect of bad data is difficult to be removed when the difference between the mean_deposit of the good agent and the malicious agent is very large. When setting the mean_deposit of the good agent as 50 and that of the malicious agent as 100, the model stability is ensured and the deposit is kept within a reasonable range.
- submission cost: The submission cost limits the frequency of submitting new data. The accuracies of trained models tend to increase with the submission cost increases. The larger the submission cost, the fewer days it will take to use up the balance of the malicious agent and the easier it will be to eliminate its negative influence to the model performance.

Because of the automated execution, privacy protection, and ability to implement distributed training, smart contract is useful to incentivize users to submit better data. According to our observation, the model accuracy improves, the gap between the accuracies before and after the simulation keeps in a relatively small range, and the balance of the bad agent can be reduced to zero in a shorter period attributed to the rules defined in the smart contract.

In the course of our work, we also notice that in existing research [11][27][29], the values of some important factors on the smart contract are picked on the basis of experience, but their impact is unknown, which is the gap we are aiming to fill.

The rest sections of this paper are organized as follows. In Section II, we give a brief introduction of the background of ML, smart contracts, and the integration of ML and blockchain; in Section III and IV, we detailed our research procedure, experiment results, and statistical analyses. We list the future directions of our work in Section V and conclude our paper in Section VI.

## II. BACKGROUND

### A. Machine Learning

ML is an evolving branch of computer algorithm in which machines emulate the way human think and learn through experience. It can be used to solve issues in diverse fields after trained with a huge amount of data, thus being an efficient tool in the era of big data. ML algorithms are useful to solve classification problems with discrete data (variables) such as classifying images and regression problems with continuous data (variables) such as forecasting temperature. The following are some classic ML models in wide use: Support Vector Machine (SVM) [12], Decision tree [13], Random forests (or random decision forests) [14], and Neural network (NN) [15] (see brief explanation in [28]). A schema of ML model training is shown in Fig. 1.

Unlike conventional ML which mostly trains models in a centralized way, CML outperforms the former through multi-party (node or agent) cooperation. Distributed ML [16][17] model training is one way of collaboration, in which different nodes co-train the model either in a data-parallel (we mainly use in this paper) or model-parallel mode [18]. Fig. 2 and Fig. 3 illustrated the visual forms of these two categories of distributed model training. In this way, CML is efficient to deal with large-scale data problems or issues that need to be solved with complex models and maintains model generalization.

Federated learning (FL) is another method for decentralized CML. Different from data-parallel training that computes with units like CPU cores or GPU devices, the FL trains the model with entities such as banks and companies, and broadcasts its updates through a centralized server. The advantages of FL are hyper-personalization, minimum latency, less consumption of power, and privacy-preserving [19]. However, it faces the risk of malicious cloud servers and unbelievable agents who may update wrongly and have a bad influence on the models [20], an issue that may be avoided or mitigated by the use of blockchain technology.

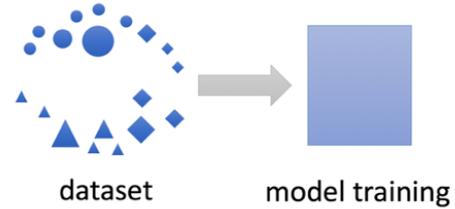

Fig. 1. Conventional machine learning model training: The dataset is centralized on one machine, which mearns the machine may have access to all sensitive private data. After exploiting the patterns of the input dataset, the machine outputs its training results. The developers then tune the model parameters through the validation of model performance before applying them to practical issues.

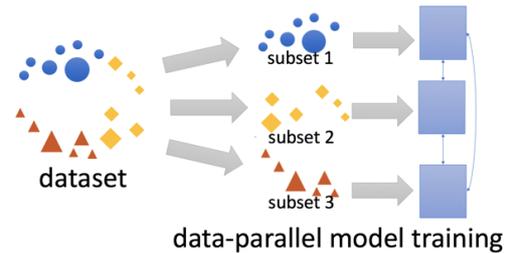

Fig. 2. Data-parallel [18] collaborative machine learning model training: In the case of large-scale data, the whole dataset cannot be stored locally due to the limitation of computer storage. The developers horizontally split the dataset into portions of subsets and then multiple nodes separately train the model using these subsets. Each node utilizes the model parameters updated by other nodes and communicates through a parameter server or peer-to-peer mode.

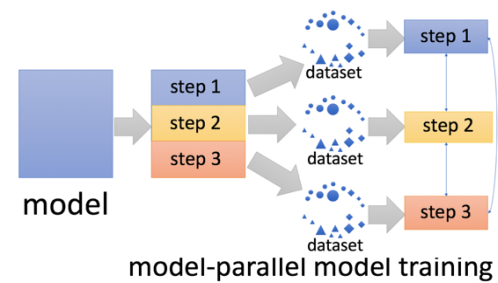

Fig. 3. Model-parallel [18] collaborative machine learning model training: In the case of complex model training, the developers deploy the model into several steps, and then each step orderly trains the copied complete dataset. The steps correspond to layers of a neural network or operation procedures in a complex formula such as addition and multiplication operations. The steps of the model utilize the model parameters updated by other steps and communicates through a parameter server or peer-to-peer mode.

## B. Smart Contracts

With the foundation of blockchain [8], a decentralized ledger that allows peer-to-peer transactions without third parties, many applications on basis of this emerging technology have appeared and offer advantages to our daily life. So far, blockchain technology has been incorporated into domains such as supply chain [21], financial services [21], smart city [22], healthcare [23], etc. Smart contract is one of the derivatives of the blockchain. The smart contract is a kind of code-form protocol on which the participants can enforce digitally defined promises. The basic idea is to embed logic terms of the contract that can automatically verify and execute in hardware and software. Ethereum [9], a successful decentralized blockchain platform, originally implements this idea of smart contracts. Fig. 4 depicts the schema of the smart contract.

Through programming a few lines of code, the smart contract can help developers realize decentralized applications and has far-reaching potential. It has been widely and successfully applied in diverse fields such as music copyright protection [24], Internet of Things [25], insurance [26], ML model marketplace [27], etc. The use of smart contract between two parties has several advantages:

- Total transparency: All the related data is truly backup on blockchain and any participant with the key can have access to the recorded and hashed data, which can be utilized for further research.

- Automated execution: There is no need for a third party to implement the protocol. The smart contract ensures automated operations of every step and guarantees the outcomes. Therefore, communication cost is reduced and miscommunication can be avoided. Automation also leads to more efficient performance.

- No paperwork: Since the smart contract is purely software and blockchain-based, paper-based contracts can be completely replaced and trust between parties is still assured. This also provides a highly efficient platform for global cooperation.

- Data privacy: In certain private chains, users without the key to the blockchain cannot have access to the data on it. Besides, the smart contract ensures data privacy between parties with the key, credit to the properties of traceability and immutability of the blockchain to a large extent.

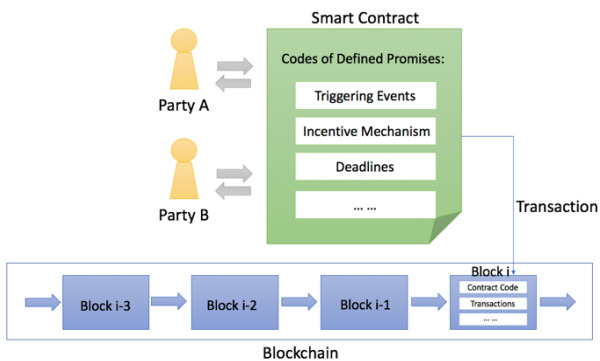

Fig. 4. A smart contract is created with written codes between two parties, with each remaining anonymous. Defined promises are set on the contract e.g. triggering events, deadlines, incentive mechanism, etc. It is then stored on a blockchain, together with the transaction data, and can verify and execute automatically.

## C. Machine Learning and Blockchain Integration

Many researchers have been conducting the study on the integration of ML and blockchain [28]. The combination of these two emerging technologies has been applied to platforms for encouraging data sharing [11], model sharing [27], or both [29]. ML and blockchain are both computer-based and require no physical connection, and therefore it is a natural choice to research their convergence, which has great potential to bring about a promising future to the development of these two fields.

The framework we mainly refer to in this paper is the Sharing Updatable Model (SUM) on blockchain [11], a platform integrating ML and blockchain for data sharing. It enables a ML model to be trained continuously with collaboratively-built datasets and utilizes smart contracts for filtering wrong or ambiguous data. The data holders uploading good data to the SUM will gain a reward and those who maliciously upload bad data will get a punishment on their deposit. Through this mechanism executed by the smart contract, the SUM is efficient in diminishing the negative influence of the bad data. The result of its simulations shows that the incentive mechanism defined on the smart contracts can successfully maintain the accuracy of the model outputs. However, we notice that the explanation of the selected values (default values) of the parameters of the incentive mechanism is missing, which may be depended on the developers' experience. Hence, we carry out a series of comparative experiments to study the relationships between variables and in-depth analyses on the effect of important factors of the smart contract and try to fill this gap on basis of this framework.

## III. RESEARCH PROCEDURES

In this section, we implement our collaborative training platform for data sharing. First, we give a detailed description of each process of this platform and how to get a reward through data sharing from the agent's perspective. Second, we specified the procedure of our simulation experiments on important factors of the smart contract defined in the platform from the developer's perspective. The model we utilize is a perceptron model for determining whether a movie review is positive or negative, and the review data used for training the model is from the IMDB reviews dataset [10].

### A. Collaborative Training Procedure

The collaborative training platform for data sharing is proposed to incentivize sharing of valuable data from diversified agents (users) for continuous training of the ML model. It allows its agents to collaboratively train the model and receive awards if the uploaded data is verified to be useful. A schematic of this platform is illustrated in Fig. 5. The following is the procedure of collaborative training on this platform:

First, an agent anonymously uploads his/her private data and deposit to the platform as required by the smart contract. It is unknown whether the agent is trustworthy at the initial stage, so a deposit (deducted from the agent's balance) is forced to store temporarily on the platform for punishment if the agent is detected malicious, or otherwise refund to the agent who is later detected reliable. Then the data is stored on the blockchain (we assume that the storage space of the blockchain is arbitrarily large) and is used to train the ML model, and the smart contract validates the data according to the output of the ML model and executes reward or punishment on the agent.

The smart contract is the core part of the platform. It is responsible for the operation of the platform on software and hardware. All the parameters and initial settings are defined on it and ensure automatic execution and verification. Important factors of the incentive mechanism include the submission cost (deducted from deposits) of the data, the size of the trained data, the value contained in the submitted data, etc. We will explore further to these in the next section. The key part of the smart contract is the incentive mechanism that verifies the validity of the submitted data and makes decisions on basis of the outcome of the trained model. For example, if a malicious agent submits wrongly labeled data, the accuracy of the model may decrease. Thereafter the incentive mechanism monitors the bad performance of the ML model and deducts part of the deposit of this agent. The deducted deposit is then used to reward the good agents or cover part of the gas fee. Besides the reward, the good agents can deposit less and wait for less period before new submission than the malicious agents. Other bonuses or restrictions can also be added to this smart contract for further use.

The other two parts of the platform are the blockchain and the ML model. The blockchain records all the related data of the platform, including the submitted data, the smart contract, the transaction data, the ML model and its results, etc. The ML model is preliminarily trained with some data and therefore has an initial accuracy. Fed with new data collaboratively gathered from various agents, the model is re-trained and obtains new training results (i.e. precision, recall, F1-scores, accuracy, etc.). With the help of the smart contract, this step is repeated and achieves continuous training.

Eventually, the agent gains the refund back after an interval (defined in the smart contract), with an increase or decrease on his/her deposit according to the validation of the smart contract.

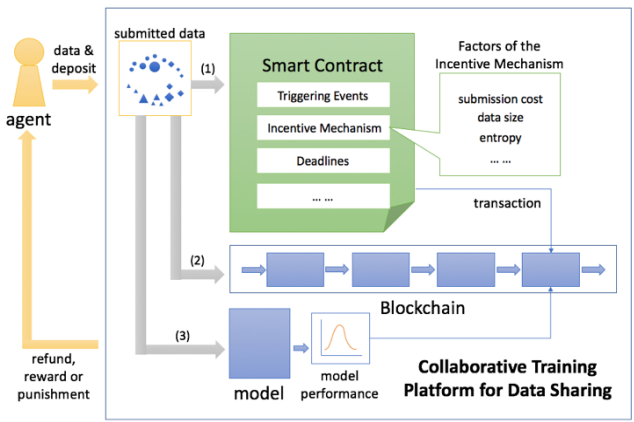

Fig. 5. Schematic of collaborative training platform for data sharing: When an agent uploads the data and deposit to the platform, the ML model will be trained with the new data. Then, the agent will get a reward or punishment, which is closely related to the model performance, and the remaining deposit (refund) computed by the incentive mechanism on the blockchain. The blockchain also records all the transaction data and the incentive mechanism defined in the smart contract.

TABLE I. SMART CONTRACT PARAMETERS, DEFINITIONS & DEFAULT VALUES

| Parameters | Definitions | Default Values | |
|---|---|---|---|
| | | Good Agent | Malicious Agent |
| num_words* | the number of binary features which presents the same number of most frequent words in the IMDB dataset | 1000 | |
| train_size* | the size of the training data of the initial model | 0.08 | |
| start_balance | the start balance of the agents | 10,000 | 10,000 |
| mean_deposit* | the mean of the deposit required to submit data | 50 | 100 |
| stdev_deposit | the standard deviation of the deposit | 10 | 3 |
| mean_update_wait_s | the mean interval (second) for the agent to wait before new submission | 10*60 | 1*60*60 |
| prob_mistake | the probability of making mistake | 0.0001 | / |
| submission cost* | the cost deducted from the deposit of the agents each time they submit new data, which is defined in the smart contract | specific value unexplained | |

*refers to most important parameters we focus on

B. Simulation Experiment Procedure

In our experiments, we utilize a perception model [30] to solve a classification problem: sentiment classification on movie reviews, with an environment set with Python (version 3.8.12), Bokeh (version 2.3.1), Keras (version 1.1.2), and TensorFlow (version 2.4.1). We reuse the code from the SUM on blockchain [11] by experimenting with different element values of the smart contract (open source provided by https://github.com/microsoft/0xDeCA10B). It is worth noting that the specific use case scenario is only for demonstration, and the model and data can be arbitrary in this research, which is of general interest.

The perceptron model is a single-layer and fully-connected neural network that can be trained with discrete data. It contains three layers: input layer, hidden layer, and output layer and needs to be trained before practical use. The input layer is fed with the training data (movie reviews data in this case). Next, its hidden layer learns the patterns from the dataset in a forward propagation way and tunes its weights and biases according to its performance through backpropagation. Then the output layer gives a judgment on the label of the input data, positive review or negative review. Finally, the trained perceptron model is verified by the test data, a dataset that is not used for training. The dataset we used for training and testing the perceptron model is the IMDB reviews dataset [10], a dataset of movie reviews from the Internet Movie Database containing a total of 25000 data samples.

The framework we mainly refered to is the SUM on blockchain [11] for simulation and validation. Specifically, SUM is a useful tool in running simulations for CML model training on blockchain. In the initial settings, a perceptron model is trained on 2000 IMDB data samples and the rest 23000 training data samples are submitted by the agents in the simulation. Each data sample includes a review (in a word vector form) and a label (positive or negative). 1000 most frequently appeared words in the reviews dataset are transferred into a dictionary (word vector) with 1000 binary features in the data samples, in which the words are its keys and the frequencies are its values.

In the simulation, two agents simultaneously share data to the platform with a deposit. The honest agent is responsible for uploading all the rest of 92% (default value that will be tuned in our experiments) of the training data samples to the model, while the malicious agent uploads disorganized or wrongly-labeled data points. According to the defined rules on the smart contract, the malicious agent is required to pay twice the deposit and wait for six times the interval before new submissions compared to the good agent. We summarize a more detailed explanation of the parameters and their default values defined in the smart contract of SUM in TABLE I. Although the SUM simulation result shows that the model performance is maintained stable despite the influence of malicious data, the effect of these parameters on the performance of the model is not known.

Therefore, we carry out a series of comparative experiments and in-depth analyses on the effect of important factors of the smart contract and try to fill this gap on basis of this framework. We also select the popular IMDB dataset and tune the values of the parameters defined on the smart contract. In general, the focus is on the starred parameters in TABLE I which is prejudged most important to the performance of the model. In each simulation experiment, only one parameter is adjusted and the result of the new model is compared to the initial one. Through our further research, these different values on the parameters do have a different impact on the performance of the model. We visualized the influence of these variables on the model in Section IV and conduct statistical analysis with our results.

## IV. RESULTS AND ANALYSES

To explore the influence of the different parameters of the smart contract on the performance of the CML model, we mainly focus on these performance metrics in our experiments:

- *Accuracy of the adjusted model*: Accuracy is one of the most important values reflecting the performance of a trained model, besides confusion matrix, precision, recall, and F1-score. It is the ratio of the number of data labeled correctly by the model to the number of total training data. In the movie review case, the dataset we used to train the model is balanced, which means the numbers of positive reviews and negative reviews are almost equivalent. Therefore, the accuracy, recall, precision, and F1-score of the adjusted model have similar values and it is reasonable to represent the performance of the model with the value of accuracy.

- *Time it costs for the balance of the malicious agent to reduce to zero*: We want to make the model robust and be invulnerable to the impact of bad data. Through monitoring how long it takes for the malicious agent to run out of his/her balance, we are able to compare the differences among parameter settings for eliminating the negative effects of bad data. If the balance of an agent becomes zero, he/she will no longer submit any data.

- *Gap between accuracies before and after simulation*: We introduce a new parameter $gap$ to measure how the adjusted model changes in its performance. It is defined in the following formula:

$$gap = accuracy_{all} - accuracy \qquad (1)$$

In equation (1), $accuracy_{all}$ means the accuracy of the model when trained with all good quality data and with no bad data. And $accuracy$ means the accuracy after simulation with a good agent and a malicious agent. The difference between the two reflects how much the performance of the adjusted model has changed from that of the original one. A negative value of $gap$ means an improvement in the model performance. We expect to have high accuracy, a short time to mitigate bad data influence, and a relatively small or negative gap in the results of our models.

### A. Influence of num_words

First, we investigate the impact of num_words, the number of the binary features in the dataset, on the model performance. Almost any ML model encludes such an important hyper-parameter, and therefore our study is of general interest. The value of num_words should be selected properly because if its value is too small, the movie review data will contain too little useful information and the experimental errors may be very large; if its value is too large, it will lead to high data storage cost, slow calculation, and long experimental cycles.

The default value of num_words is 1000. In this simulation, we separately set its new values equal to 100, 200, 300, 400, 500, 600, 700, 800, and 900 and keep the default values of all the other parameters. Fig. 6 to Fig. 8 show the influence of num_words on the gap, the accuracy of the model trained using all the data and that of the model after simulation, and the time spent to use up the balance of the malicious agent. Gaps and accuracies are measured by percentage. Time is measured by days.

Combining Fig. 6 and Fig. 7, it is clear that the accuracies tend to increase with the growth of the number of most frequent words. We expect the accuracy after simulation (red line) to be close to or even higher than the accuracy before simulation (blue line). The difference between the accuracies before and after simulation can be shown by the gap in Fig. 6. When the value of num_words is in the range from 400 to 900, the accuracies of the model before and after simulation are very close, which means the accuracy of the model after the simulation is relatively stable with the influence of the incentive mechanism, even though the malicious agents keep submitting bad data. The gap reaches its minimum value -0.42 when num_words is 800, which indicates that the accuracy after simulation (red line) is 0.42% higher than the accuracy before simulation (blue line). The model trained with the smart contract performs the best with this hyperparameter. However, when the value of num_words is larger or smaller than 800, the accucry after simulation (red line) is lower than or very close to the accuracy before simulation (blue line). When num_words is less than 400, the accuracy of the model becomes relatively unstable. This is because when the value of num_words is large enough, the complex patterns it contains enables the ML model to learn and screen bad data more correctly.

Fig. 8 shows that in general, the less the number of most frequent words is used in the dataset, the more days it takes to use up the balance of the malicious agent and the harder it is to eliminate its negative influence on the model performance. When num_words equals to 800, it takes the fewest days to use up the balance of the malicious agent. If a too-large value of num_words is chosen, the data will contain too much noise which makes it difficult to distinguish between good and bad data because of the complexity of the data and model. On the other hand, if a too-small num_words value is selected, the data will contain too little information and it will also be hard for the model to make correct judgments. Although the influence of bad data can also be decreased by the incentive mechanism when small values of num_words are selected, the model performance is undesirable. Two reasons might be can explain this:

*1)* The reduction of the value of num_words causes the loss of useful information embedded in the feature vector (word vector). Feature engineering is an important step that maps the input to a feature vector, which depicts the original features of the data. When the value of num_words is too small, which means the dimension of the feature vector is too small, the model learns so small number of features that it can hardly perform well whether trained with all data or with the help of an incentive mechanism.

*2)* When the value of num_words is too small, it is difficult for the incentive mechanism to tell the difference between good data and bad data since both lose many useful features and may not improve the performance of the model. Therefore, when the model is trained with new data, the incentive mechanism may wrongly keep the bad data. While there still exist some differences between good and bad data when the value of num_words is small, it may take a long time to filter out the bad ones. Hence the gap increases and the model performs worse after simulation. In addition to these two reasons, there could be many other reasons, e.g. data complexity.

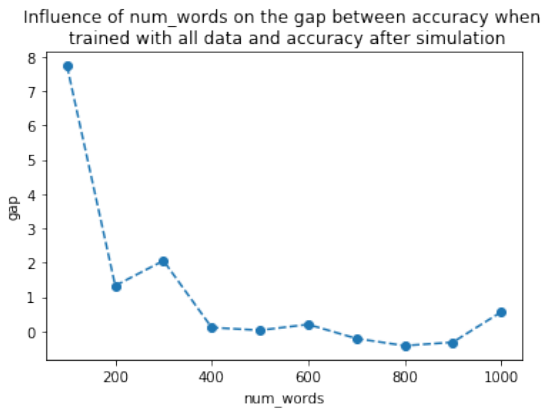

Fig. 6. Influence of num_words on gap

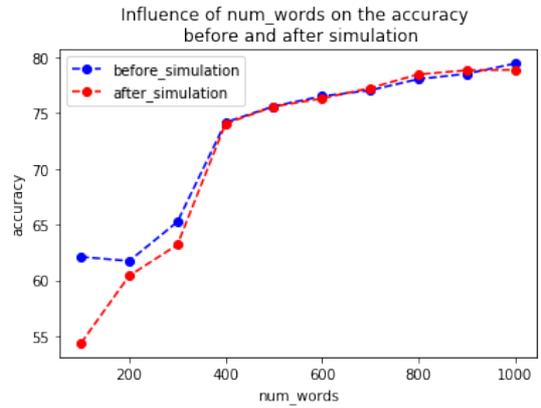

Fig. 7. Influence of num_words on the accuracy before and after simulation

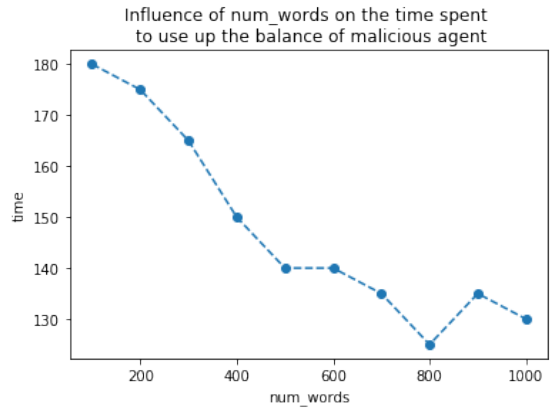

Fig. 8. Influence of num_words on the time spent to use up the balance of the malicious agent

## B. Influence of train_size

Second, we conduct multiple training experiments on train_size, the size of the training data of the model before simulation. We need an initial value of model accuracy to evaluate the performance of new models. If the train_size is too small, the issue of under-fitting may emerge and the gap cannot represent how well the new model is trained. On the other hand, if the train_size is too large, the issue of over-fitting may occur and the model may lose generalization.

The default value of train_size is 8%, which is 2000 training data out of the whole 25000 data samples. In our training experiments, we change its value into 1%, 2%, 3%, 4%, 5%, 6%, 7%, 9%, 10%, 16%, 24%, and 32%. Fig. 9 to Fig. 11 demonstrate the influence of train_size on the gap, the accuracy of the model trained using all the data and that of the model after simulation, and the time spent to use up the balance of the malicious agent. Gaps and accuracies are measured by percentage. Time is measured by days. Training step size is measured by the proportion of the whole training data set.

Comparing Fig. 9 and Fig. 10, it is apparent that the accuracies tend to increase with the mount of train_size. The difference between the accuracies before and after simulation can be shown by the gap in Fig. 9, which becomes smaller when the train_size gets larger. It is worth noting that when the value of train_size is very small, which means when too small proportion of all data is used to train the model before simulation, the accuracy is unstable and has little reference value.

A negative value of gap in Fig. 9 indicates that the accuracy after simulation is higher than the accuracy when trained with all data. The reason for this is that the larger proportion (24%, 32%) of data is set as train set for simulation, the more stable the model performance will become, compared with the initial model, which uses 8% of the data as its train set. Even though the bad data bring about negative impact to the training of the model, it can still perform better. The small fluctuations of the values of gap may be caused by randomness and in general, we expect relatively large train_size.

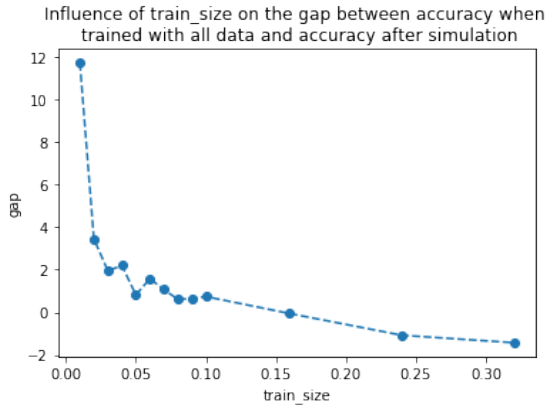

Fig. 9. Influence of train_size on gap

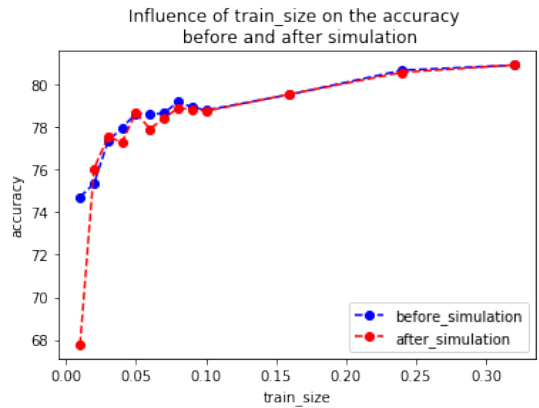

Fig. 10. Influence of train_size on the accuracy before and after simulation

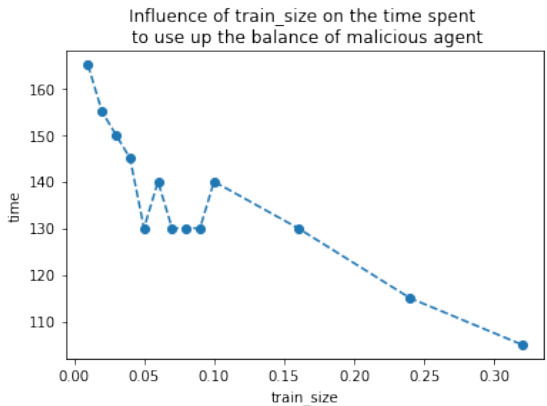

Fig. 11. Influence of train_size on the time spent to use up the balance of the malicious agent

Fig. 11 shows that the less the proportion of data set is used to train the model before simulation, the more days it will take to use up the balance of the malicious agent, and the harder it is to eliminate its negative influence on the model performance. When 32% of the data are used as training data and 68% of the data are submitted by the good agent during the simulation, the proportion of the data used for simulation gets smaller. In other words, the absolute quantity of good data submitted by the good agent declines, and so does that of bad data submitted by the malicious agent. The entire simulation period is shortened at the same time. Consequently, the negative influence of bad data can be eliminated by the incentive mechanism faster.

### C. Influence of mean_deposit

We also study the different values of mean_deposit, which is the mean of the deposit required to submit data to the platform. If the agents are to upload data freely, it would be difficult to fine a dishonest agent. Therefore, a deposit is collected in advance as the basis for incentives or sanctions on the agents. Since the mean_deposit have no impact on the training of the model before simulation, we mainly focus on its influence on the model accuracy after simulation, and the initial accuracy (blue line in Fig. 12) is only used as a benchmark for reference.

The default value of mean_deposit of a malicious agent is 100 and that of a good agent is 50. An equally proportional increase or decrease in the mean_deposit for good and malicious agent has little effect on the model, so we investigate the effect of asymmetric changes between these two parameters. We keep the value of the mean_deposit of the malicious agent as 100 and switch that of a good agent into 0.5, 5, 500, and 5000. In Fig. 12 and Fig. 13, all the values of mean_deposit are taken as logarithms with base $e$. The default values of all the other parameters stay unchanged.

Fig. 12 and Fig. 13 illustrate the influence of mean_deposit on the accuracy of the model after simulation, and the time spent to use up the balance of the malicious agent. Accuracies are measured by percentage. Time is measured by days. Cryptocurrency is used as the deposit for uploading training data.

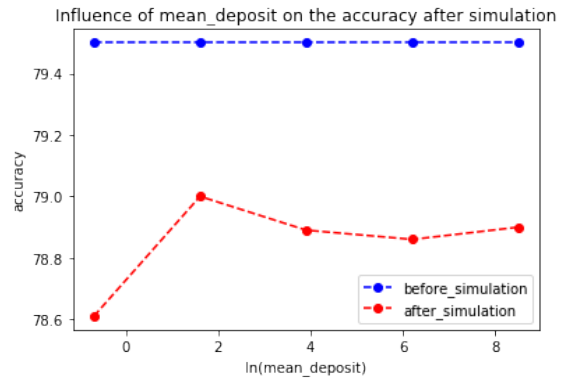

Fig. 12. Influence of mean_deposit on the accuracy after simulation

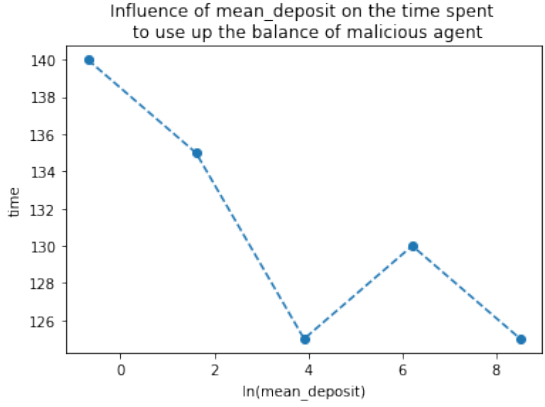

Fig. 13. Influence of mean_deposit on the time spent to use up the balance of the malicious agent.

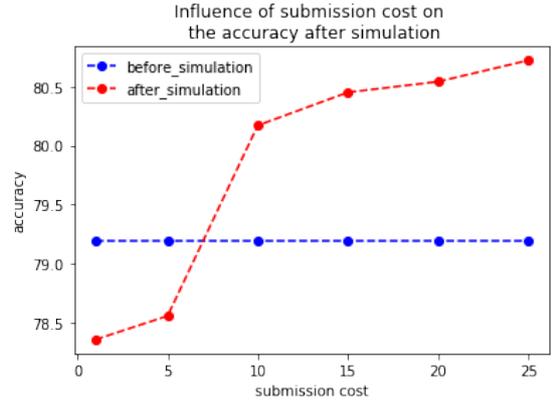

Fig. 14. Influence of submission cost on the accuracy after simulation

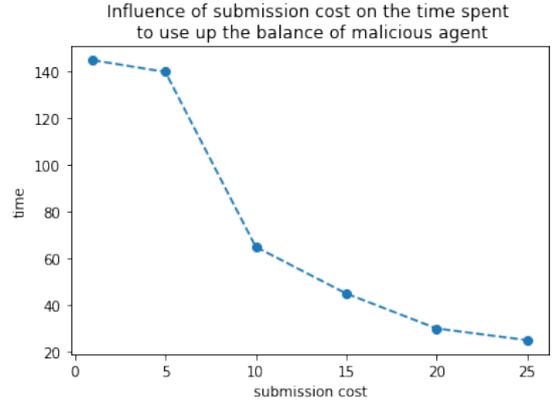

Fig. 15. Influence of submission cost on the time spent to use up the balance of the malicious agent. When the submission cost is 1, the balance of the malicious agent only decreases 2%.

As shown in Fig. 12, it is obvious that the accuracies after simulation are lower than the initial one before simulation (the red line is below the blue line), which means the effect of bad data is difficult to be removed when the difference between the mean_deposit of the good agent and the malicious agent is very large. Though tuning the deposit value of the good agent from 0.5 to 5000, the gap between the model accuracies before and after simulation is no larger than 1% (relatively smaller than the value of gap), which indicates that in general, the mean_deposit does not have as significant an impact on the overall model performance as the previous parameters (num_words and train_size). However, the mean_deposit cannot be too small (e.g. mean_deposit = 0.5), otherwise the volatility of the model results will be high.

Fig. 13 shows that when the mean_deposit of the good agent is much lower than that of the malicious agent, the balance of the malicious agent will hardly be consumed, while the mean_deposit of the malicious agent is much lower than that of the good agent, the balance of the malicious agent will be consumed more quickly. We also notice that there is one data point that ensures both model accuracy and the ability to eliminate the effects of bad data relatively quickly, that is when the mean_deposit of the good agent is 50. This is a reasonable and ideal value of mean_deposit in this case that ensures the stability of the model and keeps the deposit within a reasonable range.

### D. Influence of submission cost

Eventually, we examine the effect of the submission cost to the result of the adjusted model. It is the cost deducted from the deposit of the agents each time they submit new data to the platform. As a threshold for participation in the sharing of data, it is clear that the submission cost can influence the behavior of the agents and experiment outcome. Since the submission cost have no impact on the training of the model before simulation, we mainly focus on its influence on the model accuracy after simulation and the initial accuracy (blue line in Fig. 14) is only used as a benchmark for reference.

The exact default value of submission_cost is not directly explained by the developers. The values of 1, 2, 3, 4, 5, 10, 15, 20, and 25 are respectively selected in our comparison experiments on submission cost.

Fig. 14 and Fig. 15 illustrate the influence of submission cost on the accuracy of the model after simulation, and the time spent to use up the balance of the malicious agent. Accuracies are measured by percentage. Time is measured by days. Cryptocurrency is used as submission cost for uploading a specific sample of training data.

As shown in Fig. 14, it is obvious that the accuracies after simulation tend to increase with the submission cost increases. Fig. 15 indicates that the larger the submission cost, the less days it will take to use up the balance of the malicious agent and the easier to eliminate its negatve influence to the model performance. Therefore we can come to a conclusion that the higher the submission cost, the better the model performance and the validation efficiency. However, in practice the submission cost should not be too high, otherwise it will decrease the willingness of agents to submit new data.

### E. Influence of other parameters

We also conduct experiments for other unstarred parameters in TABLE I:

start_balance influences the model performance. The larger its value is, the slower the balance of the malicious agent is deducted to zero, which also leads to the reduced accuracy of the model. This is because when the start_balance gets larger, it spends a longer time consuming it, thus prolonging the negative influence brought about by the bad data and the performance of the model becomes unstable. However, the consumption of the start_balance is mainly affected by the values of mean_deposit and submission cost. Therefore, we classify this parameter as being of secondary importance.

mean_update_wait_s also have an impact on the model performance. The small value of mean_update_wait_s leads to frequent submissions of both good and bad data. In extreme cases, for example, thousands of data samples are submitted within an hour, the balance of the malicious agent can be deducted to zero within one day. However, the time to get a refund should be at least one week according to the rules on the smart contract, so the balance of the good agent also decreases at the initial stage, but increases obviously after a week, along with the improved accuracy of the model. While it is true that the frequency of data submission can affect the model results, there is rarely such an extreme case in practice, and agents may be unlikely to continue to participate in data sharing when their balances are used up at once with no short-term return.

stdev_deposit and prob_mistake do not have a significant impact on the model training according to our experiments, compared to the former parameters.

To sum up, the good data and bad data compete with each other during the simulation and at the same time, the incentive mechanism can reduce the negative impact caused by the malicious agent. After a period of wane and wax, the accuracy of the model keeps close to the initial one. Based on our experimental results, the balance percentage of the good agent steadily inclines and reaches 200% at the end of the simulation, while that of the malicious agent declines and becomes 0%. So we can conclude that despite the bad data, the model accuracy can still be maintained thanks to the incentive mechanism defined in the smart contract.

## V. FUTURE DIRECTIONS

To pave the way for further improvement, we list some future directions of this paper: First, the model we used for simulation is a perceptron model with only one hidden layer. It can be replaced with more complex models such as the logistic regression model or boost decision tree in further study. Second, we assume that the storage space of the blockchain is arbitrarily large. However, the blockchain storage is limited and the gas fee for holding a large amount of data and information can be very high in reality, which needs to be taken into consideration in practical operation. Third, the rules defined in the smart contract can be more diversified to ensure the quality of the data. Data quality can be measured from more various aspects [31] [32] [33]. For example, the submission cost can be linked with the entropy of the submitted data instead of setting a constant value. In this case, the better the data is, the less submission cost is required from the data holder. Whereas if an agent continues to submit bad data, he/she will face high submission fees. Last but not the least, we can find the exact reason of the gap between the accuracies before and after the simulation. We speculate that the gap may be dominated by the ratio of good data to bad data, so we can adjust that ratio in further work. The current ratio of the good and bad data is 1:1 in our research. The model can at least cancel the negative effect of the bad data, but it is not yet possible to achieve a relatively higher accuracy after training than before training.

## VI. CONCLUSION

In this paper, we have proposed a collaborative training platform that encourages valid data sharing for ML model training and maintains model robustness. Besides, it is also a secured decentralized framework that preserves the privacy of the data from various agents, credit to the use of the smart contract. Specifically, the incentive mechanism defined in the smart contract automatically verifies the data submitted by the agents, defends against malicious data sharers, and calculates the rewards or punishment for them. The blockchain keeps recording all the related data and protects data privacy.

We also carry out experiments for exploring the influence of the different parameters of the smart contract on the performance of the ML model. Three performance metrics that we mainly focus on are the accuracy of the adjusted model, the time it costs for the balance of the bad agent to reduce to zero, and the gap between the accuracies before and after simulation. From the results, we find out four most important parameters: num_words, train_size, mean_depsit, and submission cost. Generally, with the increase of the values of these four parameters, the model accuracy improves, the gap between the accuracies before and after the simulation keeps in a relatively small range, and the balance of the bad agent can be reduced to zero in a shorter period. When the value of num_words is large enough (800 in this case), the complex patterns it contains enables the ML model to learn from useful features in the data and screen bad data more correctly. When the train_size is big enough (for example larger than 5%), the accuracy of the model before simulation is stable and has a referential value for the incentive mechanism to verify the data. mean_deposit and submission cost limits the frequency the agents submit new data to the platform, and helps encourage good data sharing and eliminate the negative influence of bad data if chosen proper values. Both statistical analysis and experiment results demonstrate that this collaborative training platform for data sharing is efficient in filtering invalid data and can keep continuous training on the ML model with robustness maintained.